\documentclass[pdflatex,sn-mathphys]{sn-jnl}

\usepackage{float}
\usepackage{graphicx}
\usepackage{subfig} 
\usepackage{colortbl}

\jyear{2021}%

\theoremstyle{thmstyleone}%
\theoremstyle{thmstyletwo}%

\theoremstyle{thmstylethree}%

\begin{document}

\title[Article Title]{Undersampling and Cumulative Class Re-decision Methods to Improve Detection of Agitation in People with Dementia}


\author[1,2,4]{\fnm{Zhidong} \sur{Meng}}\email{zd.meng@mail.utoronto.ca}

\author[2,3]{\fnm{Andrea} \sur{Iaboni}}\email{andrea.iaboni@uhn.ca}

\author[4]{\fnm{Bing} \sur{Ye}}\email{bing.ye@utoronto.ca}

\author[5]{\fnm{Kristine} \sur{Newman}}\email{kristine.newman@ryerson.ca}

\author[2,4]{\fnm{Alex} \sur{Mihailidis}}\email{alex.mihailidis@utoronto.ca}

\author[1]{\fnm{Zhihong} \sur{Deng}}\email{dzh\_deng@bit.edu.cn}

\author*[2,4]{\fnm{Shehroz} \sur{S. Khan}}\email{shehroz.khan@uhn.ca}


%

\affil[1]{\orgdiv{School of Automation}, \orgname{Beijing Institute of Technology}, \orgaddress{\city{Beijing}, \postcode{100081}, \country{China}}}
\affil[2]{\orgdiv{KITE—Toronto Rehabilitation Institute}, \orgname{University Health Network}, \orgaddress{\city{Toronto}, \postcode{M5G2A2}, \state{ON}, \country{Canada}}}
\affil[3]{\orgdiv{Department of Psychiatry}, \orgname{University of Toronto}, \orgaddress{\city{Toronto}, \postcode{M5T1R8}, \state{ON}, \country{Canada}}}
\affil[4]{\orgdiv{Institute of Biomedical Engineering}, \orgname{University of Toronto}, \orgaddress{\city{Toronto}, \postcode{M5S3G9}, \state{ON}, \country{Canada}}}
\affil[5]{\orgdiv{Daphne Cockwell School of Nursing}, \orgname{Ryerson University}, \orgaddress{\city{Toronto}, \postcode{M5B1Z5}, \state{ON}, \country{Canada}}}

\abstract{Agitation is one of the most prevalent symptoms in people with dementia (PwD) that can place themselves and the caregiver's safety at risk. Developing objective agitation detection approaches is important to support health and safety of PwD living in a residential setting. In a previous study, we collected multimodal wearable sensor data from 17 participants for 600 days and developed machine learning models for detecting agitation in one-minute windows. However, there are significant limitations in the dataset, such as imbalance problem and potential imprecise labels as the occurrence of agitation is much rarer in comparison to the normal behaviours. In this paper, we first implemented different undersampling methods to eliminate the imbalance problem, and came to the conclusion that only 20\% of normal behaviour data were adequate to train a competitive agitation detection model. Then, we designed a weighted undersampling method to evaluate the manual labeling mechanism given the ambiguous time interval assumption. After that, the postprocessing method of cumulative class re-decision (CCR) was proposed based on the historical sequential information and continuity characteristic of agitation, improving the decision-making performance for the potential application of agitation detection system. The results showed that a combination of undersampling and CCR improved F1-score and other metrics to varying degrees with less training time and data.}




\maketitle

\section{Introduction}
The aging population has led to a continuous increase in the number of people with dementia  (PwD) around the world. At present, the number of PwD worldwide is about 55 million, which will double by 2050 \cite{who}. People with Alzheimer's disease and other forms of dementia often experience behavioural and psychological symptoms of dementia (BPSD) \cite{cerejeira2012behavioral}. Among them, agitation is the most common symptom, which not only increases the cost of care and manpower, but also puts the safety of PwD and caregivers at risk \cite{brodaty2014predictors}. Developing effective detection methods for agitation events can reduce the burden on the health care system and improve the quality of life for PwD and their caregivers.

Observational tools are mostly used to assess agitation e.g., PAS \cite{rosen1994pittsburgh}, and CMAI \cite{cohen1991instruction}. However, these tools are subjective and could suffer from recall bias. They cannot be used for continuous monitoring of patients and can be an overhead on the ongoing shortage of staff in dementia care facilities \cite{khan2019agitation}. Therefore, there is an urgent need to develop systems to automatically and continuously detect and report agitation events using various sensing modalities, including cameras, ambient sensors and wearable devices \cite{khan2022unsupervised,cote2021evaluation}. Among them, wearable devices with built-in multimodal sensors have essential advantages in privacy, flexibility, and cost \cite{iaboni2022wearable}. Cote et al. \cite{cote2021evaluation} used wearable technology to quantify differences in neurophysiological endpoints of PwD. It has been further demonstrated that sensor technologies have a broad range of assessing behavioural and psychological symptoms in PwD \cite{husebo2020sensing}.

Our previous work pioneered the collection of multimodal sensor data using smart wristbands from 17 PwD for 600 days, and demonstrated that leveraging the multimodal sensor data and machine learning algorithms to detect agitation was feasible \cite{spasojevic2021pilot}. However, the collected data showed that even though agitation was a clinically important symptom among PwD, normal daily behaviours still accounted for most of the time without the occurrence of agitation events. The difference between occurrence frequencies of normal and agitated behaviours led to two major limitations of the collected data. One was the severe imbalance problem that agitation (minority class) and normal behaviour (majority class) account for 1.3\% and 98.7\% of all data respectively. The other was potential imprecise labels since annotators were able to easily label the prominent stages of observed agitation, but had difficulty tracing the exact start and end times.

It is commonly agreed that sufficient data is one of the important factors to train a generalized machine learning classifier. However, the learned classifier tends to pay more attention to the majority class when encountering imbalanced data, thus reducing the overall classifier performance. We already handled this problem by weighting the minority class (i.e., agitation class) in comparison to normal activities and showed that it could improve performance in comparison to no weights \cite{spasojevic2021pilot}. However, it still does not address the problem of necessity of considering large amounts of normal data to build equally competitive classifiers to detect agitation, which is important for forwarding the future research by potentially reducing the cost of data collection and labeling. Dubey et al. \cite{dubey2014analysis} introduced different resampling methods when validating the diagnostic classifier with Alzheimer's disease images to deal with imbalanced dataset, including oversampling and undersampling approaches \cite{farquad2012preprocessing}. Oversampling increases the size of minority class to reduce the degree of imbalance, such as SMOTE \cite{chawla2002smote} and its variant ADASYN \cite{he2008adasyn}. However, generating artificial data of clinical importance (e.g., agitation) cannot be done out of the box especially for time series sensor data \cite{moniz2017resampling}. Undersampling reduces the skewed ratio of majority class to minority class by lowering the size of the former. Generally, the performances of oversampling are worse than that of undersampling with problems of overfitting and overgeneralizing when the degree of imbalance is high \cite{drummond2003c4}. It is to be noted that data collection in a real-world clinical setting is a challenging task. In the context of limited amount of agitation data, controlled reducing normal activities data is a safe alternative in comparison to collecting or artificially generating additional agitation data that is arduous and could overfit the models and would require further clinical assistance. Random undersampling (RUS) is one of the most straightforward approaches. It has been effectively applied in fields such as service level prediction\cite{pozo2021prediction}, electricity theft detection\cite{mujeeb2021electricity}, and medicare fraud detection\cite{hancock2022effects}. Saripuddin et al. \cite{saripuddin2021random} used RUS on imbalanced time series data and reported it increased true positive rate and didn't cause underfitting issues. In addition, cluster or autoencoder based undersampling methods were also used in handling imbalance problem \cite{yen2009cluster,huang2022boosting}. There are few literatures on dealing with imbalance sensor data with undersampling methods in dementia care, and we have demonstrated the evidence of their effectiveness and mitigate the potential risk of underfitting on our real-world agitation dataset in this paper. Although ground-truth class label of the dataset is another important factor in determining the performance of the classifier, it is common that only a small fraction of the collected data is actually labeled in many real-world applications \cite{vaizman2017recognizing}. We came up with a validation method to assess the quality of our dataset and showed the influence of the potential imprecise labels. Further, we investigated a postprocessing method for making better detection decisions. In this paper, three advances were made as follows:
\begin{enumerate}
    \item We showed that RUS with 20\% of the normal behaviour data produced a competitive classifier, saved the training time consumption, and reduced labeling cost of future data collection in real-world clinical settings. 
    \item We verified the reliability of our dataset, despite the possibility of existence of unseen unlabeled agitation in the transition state of agitated and normal behaviours.
    \item The proposed postprocessing  method CCR marginally improved the overall performance of the detection model trained on a smaller amount of data.
\end{enumerate}
This unique multi-modal data is available to researchers upon request and necessary ethics approvals have been obtained.

\section{Data collection and processing}\label{DCP}

\subsection{Data collection}\label{DCP1}

Our previous work completed the collection of data using the multimodal sensor framework for detecting agitation and aggression in PwD installed at a Specialized Dementia Unit at the Toronto Rehabilitation Institute (SDU-TRI), Canada \cite{khan2017daad}. The framework included wearable Empatica E4 wristbands and 15 cameras installed in public areas. We recruited 20 participants after obtaining the informed consents from their substitute decision-makers. The duration of each participant's study was set to a maximum of two months as per the ethics approval.

This paper focuses on using Empatica E4 data to detect agitation including acceleration, blood volume pulse, electrodermal activity and skin temperature. This type of multimodal data has been previously used for stress detection \cite{van2020standardized}, daily activity identification \cite{pietila2017evaluation}, and other clinical applications \cite{wang2015low, park2019smart}. In our study, Empatica E4 was put on the participant’s dominant hand in the morning and removed at night for charging and backup on the following day.

The labeling mechanism was manually implemented. Nurses in SDU-TRI were trained to record participants' behaviours and the approximate start and end times of agitation events in their nursing charts. The researchers further fine-tuned the agitation start and end using the video recordings of cameras. However, in certain cases the manual labeling mechanism would inevitably lead to inaccuracy or even lack of start and end times, which was a challenge to be evaluated in this study. These situations could happen when the PwDs were in private areas of SDU-TRI or video feeds were unavailable. Ultimately, 3 participants didn't show any agitation behaviours throughout the collection, so our studies were based on the data from 17 participants. The participants demographic information is shown in Table~\ref{tab-DPD}, where agitation day is defined as a day on which at least one agitation event occurred, and each agitation event is fully annotated with both start and end times. Similarly, a ``normal'' day is defined as a day without agitation events.
\begin{table}[htbp]
\begin{center}
\caption{Description of participants and dataset}
\begin{minipage}{200pt}
\begin{tabular}{ll}
\toprule
\# Participants                   & 17            \\ 
~~Age (years), Mean (SD)            & 78.9 (8.9)  \\
~~Age (years), Range                & 65 - 93       \\
~~\multirow{2}{*}{Gender}           & \# Males=7    \\
                                  & \# Females=10 \\ \midrule
\# Days of collection             & 600           \\
\# Agitation days                 & 239           \\
\# Fully labeled agitation events & 305           \\
\# Average of agitation duration     & 8.6 minutes     \\
~~Ratio of Normal : Agitation     & 98.7 : 1.3       \\ \botrule
\end{tabular}
\label{tab-DPD}
\end{minipage}
\end{center}
\end{table}

\subsection{Data processing and feature extraction}\label{B1}

Data processing and feature extraction adopted the same processes as described by Spasojevic et al. \cite{spasojevic2021pilot}. All sensors were resampled to 64 Hz and the high-frequency noises were removed by a first-order Butterworth low-pass filter. In this study, we used a combination of 67 features and 1-minute sliding window, which was found to be most effective window length in this dataset. The windows were non-overlapped to prevents information leakage during training and testing classifiers. To further verify the influence of the amount of data on the classifier training, the data from both agitation days and ``normal'' days were combined to form a dataset. The results by Spasojevic et al. \cite{spasojevic2021pilot} were conducted on only agitation days, whereas we showed new results on the larger combined dataset. In addition, it is worth mentioning that we found three labeling oversights from previous implementation \cite{spasojevic2021pilot}, including,
\begin{itemize}
    \item Agitations that occurred during the wristbands were being put on or taken off were all assigned normal label;
    \item Data collected earlier than 10:00 a.m. were all assigned normal label; and
    \item Agitations that lasted less than or equal to the length of sliding window were incorrectly assigned normal label.
\end{itemize}
We have made corrections so that the assigned labels are consistent with the nursing charts. The overall size of the dataset in this paper is 205,427 rows over 67 features. In the supporting supplementary document, the data sizes cross participants are reported in Table S1, and the features are listed in Table S2 along with their definitions.

\section{Methods}\label{RA}

\subsection{Undersampling methods for imbalance problem}\label{UMIP}

\subsubsection{Random Undersampling (RUS)}
Using RUS, the training data from the original majority class, i.e., normal instances, were randomly selected and added to the rebuilt training set until the preset proportion of selection was reached. The rebuilt training set retained all training data from the minority class, i.e., all agitation instances. RUS addressed imbalance problem and helped assess classifier performance under different selection proportions. This can help analyze the relation between different amounts of training data and the performance of the classifier. In addition, since the probability of each normal instance being selected is equal, RUS ensured that the distribution of the rebuilt normal set on variations of normal behaviours was consistent with that of the original normal set. That is, even if the number of normal instances decreased, the diversity of normal behaviour variations was retained. It can be illustrated by equation \eqref{pi}, where $I_{new}$ and $I_{origin}$ are the normal instances in the rebuilt training set $D_{new}$ and original dataset $D_{origin}$, respectively, and $N_{i}$ represents one of the limited normal behaviour variations.
\begin{equation}\label{pi}
    \begin{split}
        P(I_{new}\in N_{i} )&\mid _{I_{new}\in D_{new}}\\
        &\cong P(I_{origin}\in N_{i} )\mid _{I_{origin}\in D_{origin}}, \ i=1,2,3,\cdot\cdot\cdot
    \end{split}
\end{equation}

\subsubsection{Autoencoder Filter (AEF)}
An autoencoder (AE) is known to be able to detect anomalies which are expected to be unlabeled agitations in our data. We firstly trained an autoencoder with 64 hidden representations in single layer encoder and decoder. We then introduced interquartile range (IQR) analysis to filter out noises among training set. The rebuilt training set by AE Filtering (AEF) based on IQR retained all training data from the minority class. And the instances of the majority class were fed to train an autoencoder to reconstruct the input by minimizing the error between the reconstructed instances and the original instances. The absolute value of reconstruction error was taken as the score. Given the lower quartile ($Q1$), upper quartile ($Q3$) and interquartile range ($IQR=Q3-Q1$), the instance was accepted as primary normal behaviour if its score $P$ satisfied equation \eqref{iqr}.
\begin{equation}\label{iqr}
P< Q3 + k\times IQR\ \&\& \ P >  Q1-k\times IQR
\end{equation}
where $k$ is a scale related to the expected proportion of selection. This method is typically used for anomaly detection with $k = 1.5$, that is, the proportion of retained instances is $99.7\%$ when the scores are approximately Gaussian distribution. In this paper, we used AE to undersample the majority class to different proportion by setting different $k$. Unlike RUS method, there should be potential similarities in the instances rejected by AEF, and the diversity of normal behaviours in the rebuilt training set decreased compared with the original set. We can investigate the importance of the diversity of variations combining the results of RUS. Besides, AE has the ability of anomaly detection and detects unseen instances with potential inaccurate or missed labels in the normal set. If a certain $k$ results in better performance of classifier, it indicates the existence of noisy instances distributed in the original normal set.

\subsubsection{Weighted Undersampling Method (WRUS)}
\begin{figure}[htbp]
\centerline{\includegraphics[scale=0.05]{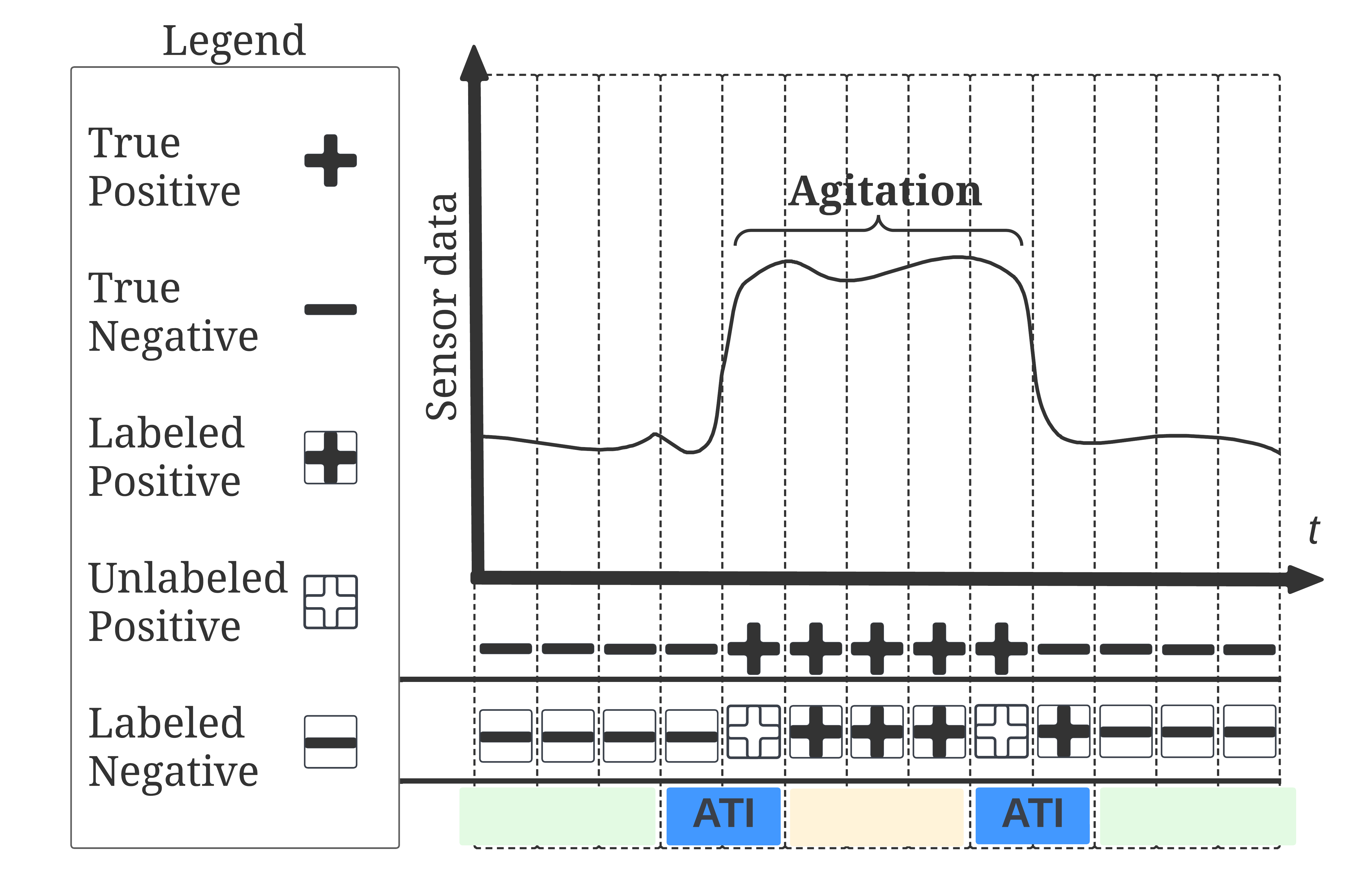}}
\caption{Illustration of ambiguous time interval.}
\label{fig-ati}
\end{figure}

As discussed in Section \ref{DCP1}, we believed that missed labeled agitation instances could potentially exist in the large-scale normal set, which can be validated by the impacts of AEF on the classifier performance. However, it was difficult to determine their distribution in the time dimension. Since our data was collected continuously and labeled by well-trained nurses, we believed the missed labels were time related and locally continuous. Gerych et al. \cite{gerych2020burstpu} modeled a similar problem as ``sequence bias'' when dealing with time series data and showed that labeled sensor data tend to occur in long contiguous sequences due to continuity of behaviour and the prevalent burst labeling manner by annotators. Similarly, we defined an ambiguous time interval (ATI) in which the instances were possible to be inaccurately labeled, as shown in Figure \ref{fig-ati}, because agitation and normal behaviour were difficult to be distinguished accurately in their transition state by human annotators. The sequence bias led to a reasonable hypothesis that inaccurate labels were most likely to occur in ATI before and after each known agitation.

\begin{figure}[htbp]
\centerline{\includegraphics[scale=0.6]{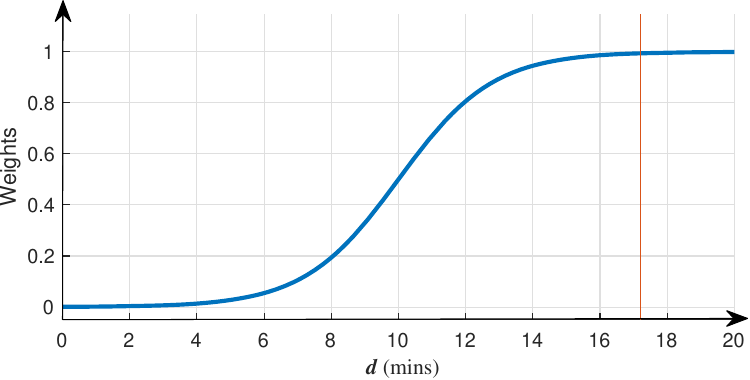}}
\caption{Weight of the probability that an instance is selected into rebuilt training set varies with time difference $d$.}
\label{fig-wrus}
\end{figure}

We proposed a weighted undersampling method based on random undersampling, WRUS, to evaluate the effects of ATI. WRUS weighted the probability that a normal instance in ATI was selected given the time difference to its closest agitation event. The calculated weights were passed into the function randsample$\left(\cdot\right)$  in MATLAB. The smaller the time difference, the lower the probability of the normal instance being added to rebuilt training set when performing undersampling. If reducing the training contribution of instances in ATI after WRUS improved the performance of the classifier, we can state with confidence that a certain amount of noisy instances existed in ATI. To this end, we deformed the sigmoid function as the weight formula as equation \eqref{sig}.

\begin{equation}\label{sig}
    w_i = \frac{1}{1+\left ( \frac{e}{\lambda_1} \right ) ^{\lambda_2\times \left ( 10-d \right ) }  } 
\end{equation}
where, $w_i$ is the weight of the $i$-th normal instance being selected, $e$ is the natural logarithm, and $d$ is the minimum time difference between the $i$-th normal instance and all agitation instances. $\lambda_1$ and $\lambda_2$ are two parameters that determine how the weight changes. Since the average duration of agitation is 8.62 minutes, we set the length of ATI twice the number, which is met by setting $\lambda_1=1.5$, $\lambda_2=1.2$. The curve of the weight formula is shown in Figure~\ref{fig-wrus}.

\subsection{Cumulative Class Re-decision (CCR)}\label{ccr}
\begin{figure}[htbp]
\centerline{\includegraphics[scale=0.6]{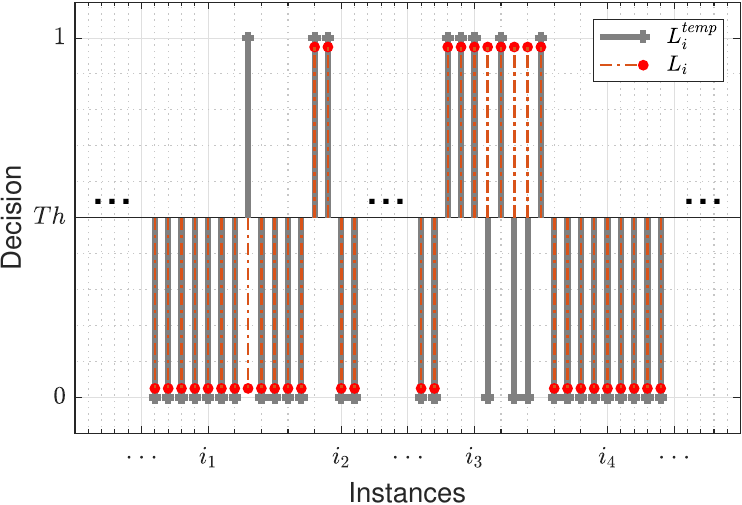}}
\caption{Illustration of CCR's effect on detection decision-making. Grey line indicates the class label directly obtained by classifier, while red dotted line is that after applying CCR.}
\label{fig-ccr}
\end{figure}

We further discussed the decision-making process of agitation detection. Compared with using the default empirical threshold ($Th$, for example: $Th=0.5$ for random forests), we explored the relationship between different thresholds and the corresponding binary classification performance. Inspired by the continuity of agitation, we proposed a postprocessing method named CCR. The class label of each instance were decided by the classifier's output and former instances in a decision window as well. The idea of CCR is illustrated in Figure~\ref{fig-ccr}. The specific rules were implemented through equations \eqref{ccr1} - \eqref{ccr3}.

\begin{equation}\label{ccr1}
    L_i^{temp}=f(S_i)=\left\{\begin{matrix}
  1 ,&S_i\ge Th \\
  0 ,&S_i <  Th
\end{matrix}\right.
\end{equation}

\begin{equation}\label{ccr2}
flag_i=\sum_{k=i-win}^{i-1}f\left (S_k\right )
\end{equation}

\begin{equation}\label{ccr3}
L_i = g\left ( flag_i  \right ) = \left\{\begin{matrix}
  0, & flag_i = 0\\
  1, & flag_i>win/2\\
  L_i^{temp}, &$otherwise$
\end{matrix}\right.
\end{equation}
where $S_i$ is classifier score of instance $i$, $f\left ( \cdot  \right ) $ is the interim decision function, $L_i^{temp}$ is the interim label of instance $i$ decided by the given threshed $Th$, $flag_i$ represents the vote of instances in the $win$-minute decision window, $L_i$ is the final label decision, $g\left ( \cdot  \right )$ is the final decision function, decided by $flag_i$. CCR firstly generates label reference of current instance based on the score of the trained classifier and takes account of the historic information of instances in the past $win$ minutes. According to the continuity characteristic of agitation, current instance is reported to be agitation when more than half of instances in the decision window are detected as agitation by the classifier. On the contrary, current instance tends to be normal if all the instances in the decision window stayed normal. Since the average duration of agitation is 8.62 minutes, the length $win$ is set to be 5 by ceiling the half of the average duration. CCR does not peek into any future information to make class decision for current instance. To the best of our knowledge, it is the first time to incorporate historical data information to develop decision-making to detect agitation using multimodal sensors.

\section{Experiment results and discussion}\label{ER}
The detection models built in this paper were all based on Random Forest with Costs (RFC), which performed better than other classifiers in our previous study \cite{spasojevic2021pilot}. An outer two-fold stratified cross-validation was used to split the dataset into primary training set and test set, each accounting for 50\% of the data. The resampling preprocessing methods were applied to the primary training set to generate undersampled training set, on which the RFC classifier were then trained. Furthermore, an inner two-fold cross-validation was used to tune the parameters of the classifier during the training process. For the parameters of RFC in MATLAB library, ``Number of Trees" was set in the range of [30,50,70,90,110]; and ``Number of Predictors to Sample" was set in the range of [1,4,7,10,...,34], which was calculated as $round\left (\frac{numf}{n}   \right )\times step+1$. In the equation, $numf=67$ refers to the number of features in the dataset; $n=20$ is a constant that determines the step size between adjacent variables in the range by $round\left (\frac{numf}{n}   \right )$, and the function $round\left ( \cdot  \right )$ performs upward rounding; and $step=0,1,2,...,\frac{n}{2}+1$ determines the total number of variables in the range. The area under the receiver operating characteristic (AUROC) was used as evaluation metric, obtained by taking the average AUROC on the test sets across the two folds. The hardware comprised of Intel Xeon E3-1240 CPU and 8GB memory. The software environment was based on MATLAB 2021b.

\subsection{Analysis on training performance under different 
undersampling proportions}\label{EA}
\begin{figure}[htbp]
\centerline{\includegraphics[scale=0.25]{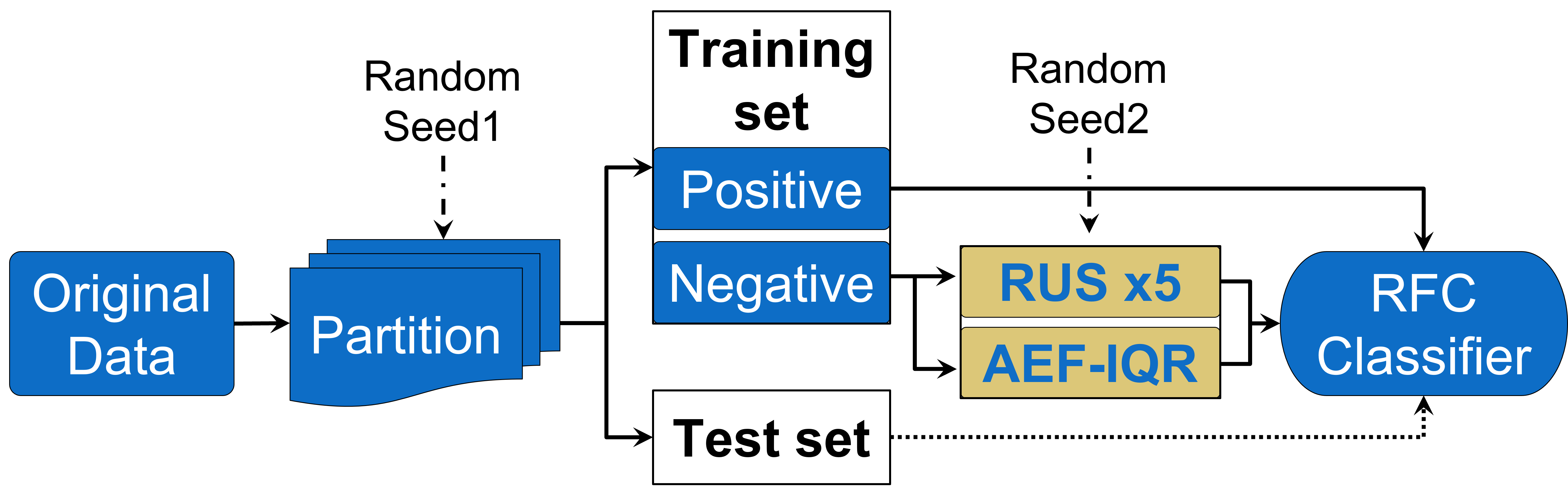}}
\caption{Flowchart of comparison experiments for RUS and AEF.}
\label{fig-rfc}
\end{figure}

\begin{figure}[htbp]
\centerline{\includegraphics[scale=0.7]{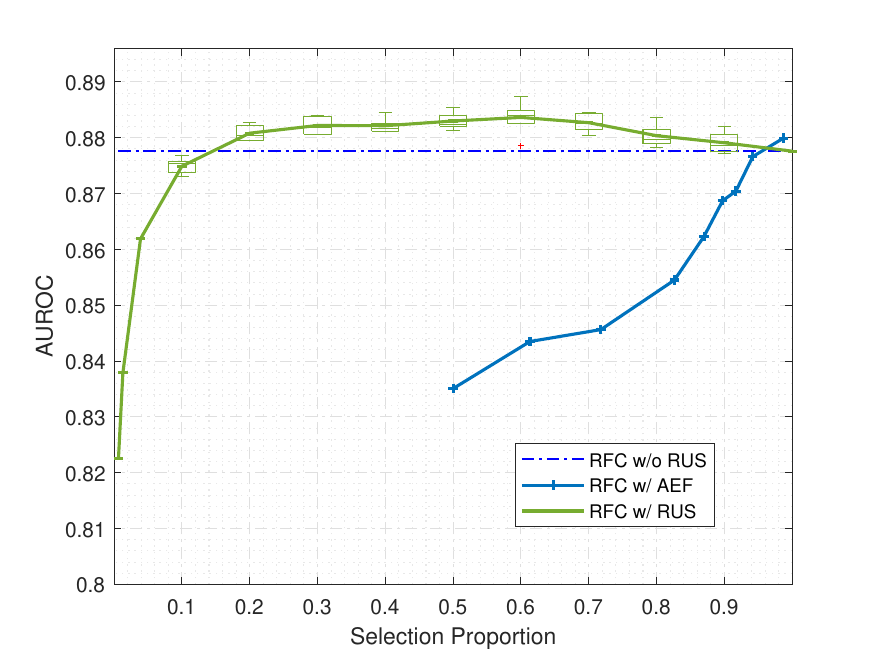}}
\caption{AUROC results of RFC without RUS, RFC with AEF, and RFC with RUS under different selection proportions; Boxplots show the IQR analysis of 5 RUS experiments.}
\label{fig-rus_aef}
\end{figure}

RUS and AEF were separately applied to generate the rebuilt training sets. RFC models were trained on the two new training sets and obtained optimal parameters respectively. Eventually, the optimal classifiers were tested on the same test set for fair comparison. The processes are shown in Figure~\ref{fig-rfc}. The proportions of selection for RUS and AEF are listed in Table~\ref{tab-sp}. Among them, 1.3\% made equal size of normal and agitation set used for training. The proportions of AEF were calculated corresponding to each $k$ in \eqref{iqr} and the AUROC is reported respectively. The random seed 1 was kept constant to ensure that the cross-validation partitions were consistent under different settings. For the RUS method, we repeated 5 times of experiments by changing Random Seed 2 and reported the average AUROC in Figure~\ref{fig-rus_aef}. The IQR analysis was shown by boxplots for better comparison when AUROC was close to the results of RFC without RUS, which was equivalent to 100\% selection proportion for RUS. The average training time is shown in Table~\ref{tab-time}.

\begin{table}[htbp]
\begin{center}
\caption{Settings of selection proportions and $k$}
\begin{tabular}{cclcc}
\toprule
\textbf{RUS}                   &  &  & \multicolumn{2}{c}{\textbf{AEF}} \\ \cmidrule{1-1} \cmidrule{4-5} 
Selection Proportions &  &  & $k$       & Proportions      \\ \cmidrule{1-5} 
0.65\%                &  &  & 0         & 50\%             \\
1.3\%                 &  &  & 0.1       & 61.3\%           \\
3.9\%                 &  &  & 0.2       & 71.8\%           \\
10\%                  &  &  & 0.5       & 82.6\%           \\
20\%                  &  &  & 1         & 86.2\%           \\
30\%                  &  &  & 1.5       & 89.7\%           \\
...                   &  &  & 2         & 91.7\%           \\
80\%                  &  &  & 3         & 94.2\%           \\
90\%                  &  &  & 10        & 98.7\%           \\ \botrule 
\end{tabular}
\label{tab-sp}
\end{center}
\end{table}

\begin{table}[htbp]
\centering
\caption{Training time consumption under different RUS selection proportions}
\label{tab-time}
\begin{tabular}{cccccccc}

\toprule
Selection proportion (\%) & 0.65  & 1.3   & 3.9   & 10    & 20    & 30    &       \\ 
Training time ($\times10^{3}s$)     & 0.170 & 0.225 & 0.396 & 0.736 & 1.201 & 1.678 &       \\ \hline
Selection proportion (\%) & 40    & 50    & 60    & 70    & 80    & 90    & 100   \\
Training time ($\times10^{3}s$)     & 2.136 & 2.577 & 3.087 & 3.512 & 3.973 & 4.430 & 4.939 \\ \botrule
\end{tabular}
\end{table}

It can be seen from Figure~\ref{fig-rus_aef} that using RFC with RUS, even if only 20\% of the normal instances were retained for training, the classifier still performed equivalently as compared to RFC without RUS. Especially when the selection proportion was between 30\% to 70\%, the performance of the classifier was even higher. The reason is that the rebuilt training set not only ensured the variation diversity of the normal activities, but also potentially mitigated the problem of imbalance to a certain extent. Table~\ref{tab-time} shows the training time of RFC with RUS is linearly related to selection proportion, indicating that an appropriate lower number of normal instances can reduce the time cost of training and improve the classifier’s performance by addressing the data imbalance problem. However, the performance of AEF is worse compared to RUS as the dotted curve is under the blue curve except when the proportion is 98.7\%. It implicitly shows that the number of mislabeled instances in our dataset is quite small. While AE tried to filter out more mislabeled instances, it eliminated normal instances with potential similarity, which reduced the diversity of normal activities and caused underfitting. In short, to obtain better performance of RFC requires adequate size of the dataset and rich diversity of instances to improve the model training. When the diversity of the dataset stays unchanged, increasing its size only increases the imbalance and time cost.

To validate the idea above, we compared the results with that of \cite{spasojevic2021pilot}, where the normal and agitation sets were obtained from agitation days only. By setting the selection proportion as 0.4 in RUS, we adjusted the size of our dataset to be the same as that of \cite{spasojevic2021pilot}, as well as the ratio of normal set size to agitation set size. We also applied RUS method to our test set ensuring the number of test instances was equal to that of \cite{spasojevic2021pilot} for comparison. The AUROC was 0.865 for \cite{spasojevic2021pilot}, while our setting increased the AUROC to 0.880. The only difference between two settings was our dataset included a richer diversity of normal behaviours from ``normal'' days, which also showed the idea that both diversity and quantity of dataset were important.

\subsection{Evaluation of the quality of dataset}\label{EB}
To further evaluate the impacts of potential mislabeled instances in ATI before and after the known agitations, we replaced the RUS with WRUS and followed the rest of experiment steps shown in Figure~\ref{fig-rfc}. The proportion of selection was set to [30\%:70\%] with separation of 10\%. The AUROC results are shown in Table~\ref{tab-wrus}. The better results in comparison are bold-faced.

\begin{table}[htbp]
\centering
\caption{AUROC of RUS and WRUS}
\label{tab-wrus}
\resizebox{\columnwidth}{!}{%
\begin{tabular}{cccccccl}
\toprule
\multicolumn{2}{c}{\multirow{2}{*}{}}     & \multicolumn{5}{c}{Selection proportions (\%)} &      \\ \cmidrule{3-7} 
\multicolumn{2}{c}{}                      & 30      & 40      & 50      & 60      & 70     & {mean} \\ \midrule
\multirow{2}{*}{AUROC of} & RUS  & 0.882        & 0.881        & 0.883        & 0.880       & \textbf{0.883}     & 0.882 \\
                          & WRUS & \textbf{0.884}        & \textbf{0.882}        & \textbf{0.885}        & \textbf{0.883}       & 0.881     & \textbf{0.883}  \\ \botrule
\end{tabular}%
}
\end{table}

Table~\ref{tab-wrus} shows that WRUS preprocessing gives slightly better results of RFC under most selection proportions compared to RUS. It indicates that there might exist a few mislabeled instances in ATI, however, the classifier is not significantly improved by reducing the contribution of those instances to training. In other word, our dataset is quite reliable for developing agitation detection algorithms.

\subsection{Assessing the performance of CCR with RUS method}\label{EC}
\begin{figure}[htbp]
\centering
\includegraphics[scale=0.7]{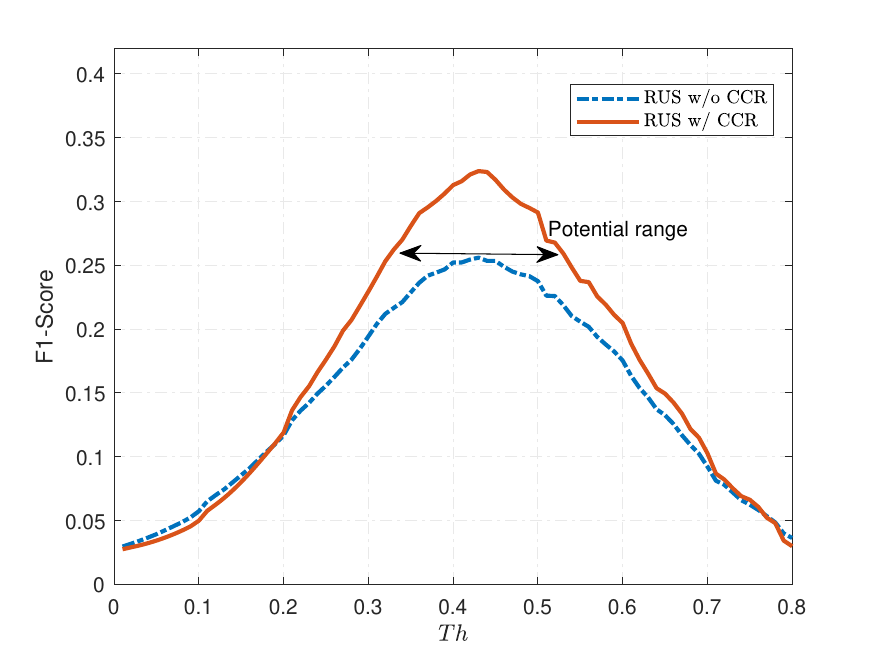}
\caption{F1-score versus different thresholds.}
\label{fig-f1}
\end{figure}

The performance of CCR method was validated based on extended RUS experiment with 20\% as selection proportion in Section \ref{EA}. To quantify the performance of decision-making, precision ($P$), recall ($R$), and F1-score ($F1$), calculated by equations \eqref{precision} - \eqref{f1score} respectively, were introduced. After obtaining the trained RFC model with RUS, we applied it on the test set and obtained the score of each instance. We firstly explored the relationship between thresholds and decision-making performance by adjusting different threshold values. Then, we followed steps \eqref{ccr1} - \eqref{ccr3} in Section \ref{ccr} to remake decision and assign class label for each instance in test set in chronological order, only relying on present and past information. It is worth noting that reporting performance on the test set is considered possible as it does not involve tuning hyperparameters or training the classifier. The curves of F1-score versus thresholds between [0.10,0.80] with separation of 0.01 are plotted in Figure~\ref{fig-f1} for both RUS without CCR and RUS with CCR methods. 

\begin{equation}\label{precision}
P=\frac{True~Positive}{True~Positive+ False~Positive } 
\end{equation}

\begin{equation}\label{recall}
R=\frac{True~Positive}{True~Positive+ False~Negative } 
\end{equation}

\begin{equation}\label{f1score}
F1=2\times\frac{ P\times R}{P + R } 
\end{equation}

\begin{table}[htbp]
\centering
\begin{minipage}{260pt}
\caption{Performance of decision-making}
\label{tab-prf}
\begin{tabular*}{\textwidth}{@{\extracolsep{\fill}}lcccccc@{\extracolsep{\fill}}}
\toprule
       & \multicolumn{3}{@{}c@{}}{RUS w/o CCR}      & \multicolumn{3}{@{}c@{}}{RUS w/ CCR}              \\ \cmidrule{2-4}\cmidrule{5-7}
$Th$   & $P$   & $R$            & $F1$           & $P$            & $R$            & $F1$           \\ \midrule
0.10   & \textbf{0.030} & 0.925          & \textbf{0.058}          & 0.026 & \textbf{0.971} & 0.050 \\
0.20   & 0.064 & 0.711          & 0.117          & 0.064 & \textbf{0.808} & \textbf{0.119} \\
0.30   & 0.125 & 0.459          & 0.196          & \textbf{0.148} & \textbf{0.532} & \textbf{0.231} \\
0.40   & 0.220 & 0.296          & 0.252          & \textbf{0.286} & \textbf{0.345} & \textbf{0.313} \\ 
0.43   & 0.258 & 0.254          & \underline{0.256}          & \textbf{0.347} & \textbf{0.304} & \underline{\textbf{0.324}} \\
0.50   & 0.340 & 0.183          & 0.238          & \textbf{0.461} & \textbf{0.213} & \textbf{0.291} \\
0.60   & 0.481 & 0.107          & 0.175          & \textbf{0.615} & \textbf{0.123} & \textbf{0.205} \\
0.70   & 0.627 & 0.050          & 0.092          & \textbf{0.751} & \textbf{0.055} & \textbf{0.103}          \\
0.80   & 0.721 &\textbf{ 0.019}          & \textbf{0.036}          & \textbf{0.816} & 0.015 & 0.030          \\ \botrule
\end{tabular*}
\end{minipage}
\end{table}


Table~\ref{tab-prf} shows part of metrics when the threshold is between [0.10,0.80] with separation of 0.10, while 0.43 is the threshold that gives the best F1-score for both RUS with CCR and without CCR. The best F1-score after applying CCR increases by 26.6\% comparing $F1$ with underlines in Table~\ref{tab-prf}. Figure~\ref{fig-f1} shows the curves of F1-score first rise and then drop after their peaks as threshold increases. It means that having high precision and recall values at the same time is a challenge in this application because a larger threshold makes more instances classified as normal by the decision equation~\eqref{ccr1}, resulting in smaller false positive (FP) and larger false negative (FN). However, these results should be treated with caution as the overall F1-score obtained with CCR are still deemed low to be used in clinical practice. The low precision values at optimal threshold indicated the presence of large amounts of false positives. A clinical evaluation may be necessary to understand these false positives and how they can impact the clinical burden.

\section{Conclusions and Future Directions}\label{CC}
Developing objective agitation detection system is valuable and promising by utilizing wearable sensors in clinical settings, which currently faces the problem of data imbalance and insufficient reliable data. This study first reconstructed a new multimodal sensor dataset for agitation detection in PwD, which covered more diverse “normal” data compared to previous study. A WRUS method inspired by ATI and manual labeling mechanism was proposed to evaluate the quality of the dataset. In addition, RUS and AEF were compared to analyze the necessity and effectiveness of addressing imbalance problem. In the situation where the RFC model faced a bottleneck in performance improvement, CCR method was proposed to promote decision-making performance based on the scores and historic information in decision windows in terms of the continuity characteristic of agitation.

The reconstructed dataset was used to validate proposed methods. It was shown by comparing results of RUS and AEF methods that the diversity of data is meaningful for training a classifier. Simply increasing the data quantity has little positive impact, and could lead to redundancy, imbalance, and time consumption for building the model. In addition, the equivalent level of results of WRUS indicated that the dataset is reliable and dependable on developing agitation related analysis, even though manual annotator may have misreported few agitations in ATI during data collection. The CCR method that was independent of training process slightly improved the decision performance of agitation detection on the basis of RUS under a reasonable threshold range, while requiring less normal data and less time. It showed that post-processing can be an effective idea to make up for the low classifier performance.

However, there are currently limitations in promoting practical applications. The achieved performance after CCR method may still be deemed low to adopt the system in a clinical care setting. The major reason is the presence of false positive in the decision making. A study is needed to understand the impact of false positives that may be accepted in a clinical environment without affecting the clinical burden by involving clinicians. Even though the amount of normal data required for training can be significantly reduced, the agitation date still needs to be accurately labeled, which poses a challenge to develop active learning approaches \cite{bianco2023reducing} for labeling where clinical experts can provide feedback on relabeling the false positives to improve overall performance.

In the future, we will deploy the agitation detection system in a real-world setting at SDU-TRI and explore effective online data annotation schemes. We will also explore the use of sequential deep learning approaches such as long short-term memory and temporal convolution networks \cite{shojaedini2020mobile,gopali2021comparison}. The necessary ethics approval has been obtained and our multimodal data will be available upon request.





\section*{Statements and Declarations}

\begin{itemize}
\item Funding. Alzheimer’s Association Research Grant.

\item Competing interests. The authors declare that they have no competing interests.

\item Ethics approval. This work involved human subjects in its research. Approval of all ethical and experimental procedures and protocols was granted by UHN-REB under Approval No. 14-8483.

\item Consent to participate. Substitute decision-makers provided written consent on behalf of the PwD. The staff also provided written consent for sensor recording in the unit.

\item Consent to publish. Not applicable.

\end{itemize}

\bibliography{nnRefs}


\end{document}